\documentclass[pdflatex,sn-mathphys]{sn-jnl}

\usepackage{lmodern}
\usepackage{lscape}
\usepackage{rotating}
\usepackage{graphicx}
\usepackage[normalem]{ulem}

\jyear{2021}%

\theoremstyle{thmstyleone}%
\theoremstyle{thmstyletwo}%

\theoremstyle{thmstylethree}%

\begin{document}

\title[Perception of Industrial Robot Movements]{The Influence Of Demographic Variation On The Perception Of Industrial Robot Movements}


\author*[1]{\fnm{Damian} \sur{Hostettler}}\email{damian.hostettler@student.unisg.ch}

\affil*[1]{\orgdiv{Institute of Computer Science}, \orgname{University of St. Gallen}, \orgaddress{\street{Rosenbergstrasse 30}, \city{9000 St. Gallen}, \country{Switzerland}}}


\abstract{The influence of individual differences on the perception and evaluation of interactions with robots has been researched for decades. Some human demographic characteristics have been shown to affect how individuals perceive interactions with robots. Still, it is to-date not clear whether, which and to what extent individual differences influence how we perceive robots, and even less is known about human factors and their effect on the perception of robot \emph{movements}. In addition, most results on the relevance of individual differences investigate human-robot interactions with humanoid or social robots whereas interactions with industrial robots are underrepresented. We present a literature review on the relationship of robot movements and the influence of demographic variation. Our review reveals a limited comparability of existing findings due to a lack of standardized robot manipulations, various dependent variables used and differing experimental setups including different robot types. In addition, most studies have insufficient sample sizes to derive generalizable results. To overcome these shortcomings, we report the results from a Web-based experiment with 930 participants that studies the effect of demographic characteristics on the evaluation of movement behaviors of an articulated robot arm. Our findings demonstrate that most participants prefer an approach from the side, a large movement range, conventional numbers of rotations, smooth movements and neither fast nor slow movement speeds. Regarding individual differences, most of these preferences are robust to demographic variation, and only gender and age was found to cause slight preference differences between slow and fast movements. }




\maketitle

\section{Introduction}\label{sec1}
From personality research in psychology, to personalized advertisements in marketing, and technical implementations such as user-specific, adaptive systems and interfaces in the field of human-computer interaction (HCI), better understanding how and why individuals react differently to certain stimuli or circumstances has been a research subject across a wide variety of scientific disciplines. In the human-robot interaction (HRI) field, a large number of studies investigate the effect of individual differences on the perception and evaluation of interactions with robots. This includes research on how demographic user profiles, but also dynamic behavioral and physiological user data, can be used to create favorable outcomes such as increased acceptance~\cite{Syrdal}, psychological safety~\cite{Lasota}, trust~\cite{Story2022} or reduced cognitive workload~\cite{Story2022,HostettlerBektas} towards improved collaborations with robots. Current technologies (e.g., contextual or on-body sensors as well as head-mounted devices) even enable the sensing of user states in real time, and thereby permit the adaption of robot behaviors to increase the acceptance of these devices at run time. 

Even though some studies have described the usage of such technologies to adapt robot behavior to user profiles~\cite{ROSSI20173, Umbrico, Kothig}, the vast majority of existing studies consider static demographic user data (see Table~\ref{LiteratureReview}). Still, the encountered effects are inconclusive and inconsistent, and the experimental settings differ regarding robot types, tasks, and samples investigated. Regarding manipulations of robot-related factors that affect human perception, many studies mix behavioral features and appearance, and relate to specific contexts, hindering generalization. Moreover, most studies use humanoid or social robots, leading to an underrepresentation of industrial robots installed in manufacturing shopfloors despite their high relevance---the continually rising number of collaborative robots (cobots) increases the number of humans who closely collaborate with industrial robots even more, and hence calls for more prominent consideration of industrial robots in HRI. 

With our study, we investigate the effect of demographic user data on the perception of industrial robot \emph{movements}, and focus on movement variations of articulated robots that have previously been shown to affect human perception~\cite{Hostettler}. To present the background of our study, the following Section~\ref{sec2} covers current relevant findings on the effect and relevance of individual differences in HRI, and systematically surveys the literature to contextualize results from ten papers that investigated the influence of individual differences on robot movement perception. Section~\ref{hypotheses} then explains how we link demographic characteristics to preferences for specifications of five MoveTypes: approach direction, movement range, rotations, smoothness, and speed. This forms the foundation for a Web-based experiment that we describe in Section~\ref{Methods}. We present the results of this experiment in Section~\ref{Results}, and discuss and compare our findings with the existing state of research in Section~\ref{Discussion}. The paper concludes with limitations and an outlook on future research directions in Section~\ref{Conclusion}.

\section{Related Work}\label{sec2}

Early investigations in the late 1800s and early 1900s tried to explain correlations such as the relation between individual mental abilities and task performance~\cite{spearman, DILLON1996619}. In the marketing field, basic demographic criteria such as age, gender, or income have been used extensively to segment customer clusters that allow customized development, marketing, and selling of products and services~\cite{Wendell, Foedermayr}. Findings from differential psychology proposed metrics like cognitive abilities, personality traits, or work performance which have been included in many HCI studies with a view to optimize user-specific experience, task performance, and technology acceptance~\cite{DILLON1996619}. Already in 1996,~\cite{DILLON1996619} observed slow research progress in characterizing user groups and that high context sensitivity complicates predictive power regarding response to and acceptance of new technologies.

In the HRI field, similar objectives have put forth findings on how user-specific characteristics can be used to optimize interactions with robots, pursuing overall goals of robot acceptance, psychological safety, cognitive workload, and improved task performance~\cite{Liu}. Individual differences in HRI research embrace several internal factors of individuals that potentially influence how humans perceive and evaluate interactions with robots. The field's breadth is reflected by the variety of names for this, and adjacent, concepts, such as \emph{individual differences}, \emph{human factors}, \emph{user characteristics}, \emph{user profiling}, etc. There is a long record of research on the general effect of individual differences on interactions among humans, such as the influence of \emph{gender} on the reaction to \emph{personal space intrusion}~\cite{Krail}, that inform HRI researchers and practitioners about potential moderating effects. To name some some of these findings that are particularly relevant to our work, \cite{Kuo} investigated the influence of \emph{age} and \emph{gender} with an ActivMedia Robotics \emph{Peoplebot} in a healthcare setting, and finds only limited relevance of age, but males having a significantly more \emph{positive attitude} towards robots than females. \cite{wagner} confirms a more positive attitude of males but found \emph{females} to assess the studied interactions \emph{more useful and satisfying}. \cite{Graaf} reviews additional studies and reports that \emph{older} people are more likely to \emph{enjoy using} and to \emph{anthropomorphize} service robots, while showing more \emph{negative emotions} towards them and \emph{lower intention to use}. \cite{Arras} finds that gender affects human attitude and anxiety towards robots in general, where males seem to perceive robots to be \emph{more useful} and are \emph{more willing to accept} robots compared to females; contrary findings have been published~\cite{nomura}. Also, culture~\cite{Kaplan} and nationality~\cite{Graaf} seem to influence the evaluation of interactions with robots. On the other hand, several studies found \emph{no significance} of individual differences, such as irrelevance of \emph{age} and \emph{prior experience with robots} on perceived human-likeness when observing an industrial robot~\cite{Hostettler} or similar occurrence of \emph{similarity- and complementarity-attraction} between participants' and the robot's personality traits using a Pepper\footnote{\url{https://www.aldebaran.com/en/pepper}} robot. In addition, similarity-attraction was more often observed with traits related to competence and intellectual abilities, whereas traits linked to social skills tend to relate to complementarity-attraction.~\cite{Craenen}.

This selection of studies demonstrates that some demographic characteristics have been found to influence individuals' perception of robots, but that most of the relationships remain unclear. The inconsistency of experimental designs, sample structures, contexts, and used robot types creates a further barrier to generalization. Regarding robot types, most of the existing findings concern social or humanoid robots rather than industrial robots, and the relationship between individual differences and specific industrial robot movements has only received little attention so far. Our goal is therefore to link industrial robot behavior to human evaluations while considering individual differences. Even though a growing number of studies investigates relationships between robots and human perception, they are mostly incapable of supporting us in generating hypotheses as they have not explicitly analysed the influence of individual differences or robot movements. For example,~\cite{castro-gonzalez} demonstrates relevant effects of smooth vs. mechanistic movements without considering individual differences, and~\cite{Hostettler} found effects of movement characteristics on perceived human-likeness but did not find an influence of participants' demographics. \cite{Ivanov2018ConsumersEstablishmentse}, on the other hand, found age and gender differences for services performed by robots in hotels, and \cite{SZCZEPANOWSKI2020100521, Dinet2014ExploratoryUsers} found education, gender, and age biases based on a comparison of different robot appearances, with both not considering the effects of robot movements. To overcome these shortages, we conducted a literature review with the goal to systematically generate links between robot movements and human perceptions, including the influence of individual differences, and explain our review approach in the following Section \ref{review}.

\subsection{Systematic Literature Review}\label{review}

To systematically review existing studies, we apply a focused approach searching scientific literature's abstracts for several criteria derived from our research goal. With regard to study types, we focus on empirical HRI research and apply the keywords [``HRI'' OR ``Human-Robot Interaction''] AND [``Empiric'' OR ``Empirical'' OR ``Experiment''] to our advanced Scopus search. While we are interested in movements of industrial robots that do not have human-like features such as eyes or a voice, in this way we do not restrict our review to only industrial robots---this is important since findings on humanoid robots' behaviors often include valuable findings on certain movements' effects, such as gestures, that are applicable more widely. On the other hand, an industrial robot's expressive capacities are limited. We therefore additionally apply the following keywords [``Movement'' OR ``Motion'' OR ``Speed'' OR ``Proxemics'' OR ``Spatial'' OR ``Angle'' OR ``Gesture''], which are derived from existing literature in the topic area. Last, to include individual differences, we explicitly focus on basic demographic criteria, and purposely do not consider complex characteristics such as cognitive abilities or personality. We therefore include [``Age'' OR ``Gender'' OR ``Education'' OR ``Experience'' OR ``Familiarity'' OR ``Location'' OR ``Nationality'' OR ``Ethnicity'' OR ``Culture''] as keywords. Also, we only include publications written in English. These search criteria were applied to Scopus document search resulting in 137 publications. Reading of the abstracts of these 137 articles leads to exclusion of 108 articles, mainly because they focused on movement characteristics of the human rather than the robot (34 publications), modulation of robot differences instead of inclusion of human differences (19 publications), or specific applications that cannot be transferred to industrial robot movements (42 publications). In the next step, the remaining 29 articles were read to ensure appropriate foci of the studies. From these 29 studies, another 19 have been excluded because they focus on specific movement manipulations (or combinations of movements) that are related to humanoid features and cannot be implemented with industrial robots or other more articulated robot types such as drones or medical robots.
Finally, 10 articles were considered for the analysis. Four of the publications include, in addition to demographic characteristics, dimensions that we do not consider, such as personality traits, verbal communication, or facial expression. These are hence still valuable to reveal findings on certain movements that can be approximated with an industrial robot. In addition, some of the included publications map movement perceptions to perceived human-likeness or anthropomorphism.

\begin{sidewaystable}[ht]
\caption{Literature Review Overview}
\label{LiteratureReview}
\resizebox{0.9\textwidth}{!}{%
\begin{tabular}{lllllllll}
\textbf{Number} & \textbf{Publication}                               & \textbf{Demographic Criteria}                                                                                     & \textbf{Movement \& robot types}                                                                                                           & \textbf{Type of interaction} & \textbf{Movement specification}                                                                                                                                                                                                                                                                                               & \textbf{Sample size \& structure}                                                                                                                                                                                                                                                                                                                                                                                                               & \textbf{General effects}                                                                                                                                                                                                                                                                                                                                                                                                                                                    & \textbf{Effects regarding individual differences}                                                                                                                                                                                                                                                                                                                                                                                                                                          \\ \hline
1               & \cite{Syrdal}                     & \begin{tabular}[c]{@{}l@{}}-Gender\\ -Personality traits\end{tabular}                                             & \begin{tabular}[c]{@{}l@{}}-Proxemics   \\ -Approach direction\\ -Peoplebot\end{tabular}                                                   & Physical Interaction         & \begin{tabular}[c]{@{}l@{}}-Robot approaches directly \\ from the front\\ -Robot approaches from the \\ front right of the participant\end{tabular}                                                                                                                                                                           & \begin{tabular}[c]{@{}l@{}}-n=33\\ -University staff and students \\ -Age range from 18 to 50, \\ median = 23\\ -20 males, 13 females\end{tabular}                                                                                                                                                                                                                                                                                              & \begin{tabular}[c]{@{}l@{}}-Participants allowed the mechanical robots \\ to approach to a closer distance than the \\ humanoid robots\\ -Participants preferred the robot to approach \\ closer for physical interaction than for verbal \\ or no interaction\\ -Participants allowed the robot to approach \\ closer from the side rather than directly from the front\\ -Habituation effect: participants allowed the \\ robot to approach closer in week 5\end{tabular} & \begin{tabular}[c]{@{}l@{}}-No effects in approach distance preference \\ for age, academic background or computer \\ proficiency\\ -Females allowed the robot to approach closer \\ than males when it was approaching directly \\ from the front\\ -No difference between preferred approach \\ directions for females, whereas males allow \\ the robot to approach closer when the robot \\ approaches from the side rather than directly \\ from the front\end{tabular}               \\ \hline
2               & \cite{Leearticle}                 & \begin{tabular}[c]{@{}l@{}}-Gender\\ -Personality traits\end{tabular}                                             & \begin{tabular}[c]{@{}l@{}}-Moving  angles\\ -Moving speed\\ -Autonomous movements\\ -AIBO\end{tabular}                                    & Physical Interaction         & \begin{tabular}[c]{@{}l@{}}-Motion angles: two to three times \\ wider in the extroverted condition\\ -Motion speed: faster in the \\ extroverted condition, 50-70\% shorter \\ interpolation between motions\\ -Autonomous movements: AIBO \\ walks and wags ist tail when not \\ responding to users' commands\end{tabular} & \begin{tabular}[c]{@{}l@{}}-n=48\\ -48 out of 200 students with the \\ most consistent scores on personality \\ tests\\ -24 extrovert and 24 introvert\\ -Only native English speakers\\ -Age range from 19 to 34, \\ median = 22.46\\ -Gender approximately balanced\end{tabular}                                                                                                                                                              & \begin{tabular}[c]{@{}l@{}}-Participants recognized AIBO's personality\\ -People applied a consistent social rule to AIBO\end{tabular}                                                                                                                                                                                                                                                                                                                                      & \begin{tabular}[c]{@{}l@{}}-Complementarity attraction effect in all \\ dependent variables\\ -Introverted people considered the extroverted \\ AIBO more intelligent, and vice versa\\ -Complementarity attraction effect on social \\ attraction, enjoyment of interaction and social \\ presence\\ -Mediating role of social presence in people's \\ social responses towards AIBO \\ -Gender was not a significant covariate for all \\ dependent and mediating variables\end{tabular} \\ \hline
3               & \cite{Story2022}                  & \begin{tabular}[c]{@{}l@{}}-Experience with an \\ industrial robot arm\end{tabular}                               & \begin{tabular}[c]{@{}l@{}}-Speed\\ -Proxemics\\ -UR5\end{tabular}                                                                         & Physical Collaboration       & \begin{tabular}[c]{@{}l@{}}-Three speed settings: 60\%, 80\% \\ and 100\% of the robot's maximum \\ speed \\ -Two proximity settings: 0.2m and\\  0.3m\end{tabular}                                                                                                                                                           & \begin{tabular}[c]{@{}l@{}}-n=83\\ -40 students and staff\\ -Age range from 21 to 53, \\ median = 25\\ -25 male, 15 female\\ -43 participants from an Academy \\ for Skills \& Knowledge and BAE Systems\\ -Age range from 16 to 59, median = 19\\ -35 male, 8 female\\ -Different levels of frequency of experience \\ with an industrial robot arm,   approx. 60\% \\ without experience, approx. 25\% with \\ weekly experience\end{tabular} & \begin{tabular}[c]{@{}l@{}}-Significant relationship between workload and \\ speed (60\% vs. 100\%), but not for proximity\\ -No significant relationship between trust and \\ speed nor with proximity\end{tabular}                                                                                                                                                                                                                                                        & \begin{tabular}[c]{@{}l@{}}-Experience of participants revealed no \\ significant relationship between experience \\ and workload nor between experience and \\ trust\end{tabular}                                                                                                                                                                                                                                                                                                         \\ \hline
4               & \cite{Bishop2019SocialAcceptance} & \begin{tabular}[c]{@{}l@{}}-Age\\ -Gender\\ -Education\\ -Experience with \\ technology\\ -User mood\end{tabular} & \begin{tabular}[c]{@{}l@{}}-Extreme vs. small movements\\ -Pepper\end{tabular}                                                             & Online Observation           & \begin{tabular}[c]{@{}l@{}}-Happy behavior: extreme \\ movements\\ -Sad behavior: small movements\\ -Neutral: an average of the two\end{tabular}                                                                                                                                                                              & \begin{tabular}[c]{@{}l@{}}-n=86\\ -Student participant pool and social media\\ -Age range from 18 to 72, median = 29.26\\ -20 male, 66 female\\ -71 of the 86 participants well educated \\ (studying or undergraduate degree and above)\\ -39.5\% somewhat familiar and 32.6\% not \\ very familiar\end{tabular}                                                                                                                              & \begin{tabular}[c]{@{}l@{}}-The neutral robot was perceived less helpful \\ than the positive robot\\ -The positive robot was perceived more foolish \\ than both, the neutral and the negative robot\\ -Sad/lonely participants were more dependent \\ on others when deciding whether to use the robot\end{tabular}                                                                                                                                                       & \begin{tabular}[c]{@{}l@{}}-Negative correlations between age and \\ perceived enjoyment and social influence\\ -No significant relationship between gender \\ and acceptance\\ -No significant relationship between education \\ and acceptance\\ -Participants more familiar with robots felt \\ they had less sufficient knowledge to use the \\ system and found the robot less safe\\ -Familiarity negatively correlated with robot \\ likeability\end{tabular}                       \\ \hline
5               & \cite{Vannucci}                   & -Culture/Nationality                                                                                              & \begin{tabular}[c]{@{}l@{}}-Speed\\ -iCub\end{tabular}                                                                                     & Physical Collaboration       & \begin{tabular}[c]{@{}l@{}}-Three speed profiles: slow, \\ medium, fast\\ -Movements inspired by \\ biological, human-like movements\end{tabular}                                                                                                                                                                             & \begin{tabular}[c]{@{}l@{}}-n=23\\ -Students, lab technicians and administrative \\ staff\\ -15 Italian and 8 Japanese participants\\ -Mean age in Italy = 30, 6 males, 9 females\\ -Mean age in Japan = 29 years, 3 males, \\ 5 females\end{tabular}                                                                                                                                                                                           & \begin{tabular}[c]{@{}l@{}}-Human movement speed was adapted to robot \\ speed and task conditions did not change \\ the behavior in Italy\\ -Human movement speed was adapted to robot \\ speed and task conditions did not change the \\ behavior in Japan\end{tabular}                                                                                                                                                                                                   & \begin{tabular}[c]{@{}l@{}}-No significant difference regarding adaption \\ of speed between Italy and Japan\\ -No significant difference regarding shared \\ personal space between Italy and Japan\end{tabular}                                                                                                                                                                                                                                                                          \\ \hline
6               & \cite{Abel}                       & -Gender                                                                                                           & \begin{tabular}[c]{@{}l@{}}-Human-like vs. \\ robotic movements\\ -Virtual gantry robot\end{tabular}                                       & Online Observation           & \begin{tabular}[c]{@{}l@{}}-Human-like movement: digressive \\ (concave) curve\\ -Robotic movement: point-to-point \\ movements, progressive (convex) \\ curve\end{tabular}                                                                                                                                                   & \begin{tabular}[c]{@{}l@{}}-n=40\\ -Participants recruited using university \\ blackboards, mailing lists and word of mouth\\ -Age range from 18 to 45\\ -20 male, 20 female\\ -Balanced samples within gender groups \\ regarding age, occupations, and education\end{tabular}                                                                                                                                                                 & \begin{tabular}[c]{@{}l@{}}-The human model was not perceived more \\ anthropomorphic than the robot model\\ -The movement mapped from human kinematics \\ was perceived more anthropomorphic than the \\ robotic movement\end{tabular}                                                                                                                                                                                                                                     & \begin{tabular}[c]{@{}l@{}}-Males were sensitive to robotic vs. \\ anthropomorphic movement differences\\ -Females showed no difference between \\ movements\\ -Females attributed more anthropomorphic \\ features to robotic movements\\ -No significant interaction between \\ movement and model\end{tabular}                                                                                                                                                                          \\ \hline
7               & \cite{Hostettler}                 & \begin{tabular}[c]{@{}l@{}}-Gender\\ -Age\\ -Location\\ -Education\\ -Prior experience\end{tabular}               & \begin{tabular}[c]{@{}l@{}}-Speed\\ -Part approach\\ -Smoothness\\ -Rotation\\ -Movement range\\ -Approach Direction\\ -UR10e\end{tabular} & Online Observation           & \begin{tabular}[c]{@{}l@{}}-Speed: slow vs. fast\\ -Part approach: with care vs. \\ without care\\ -Smoothness:  calm vs. \\ nervous\\ -Rotation: conventional vs. \\ unconventional\\ -Movement range: small vs. large\\ -Approach direction: from the side \\ vs. straight towards participant\end{tabular}                 & \begin{tabular}[c]{@{}l@{}}-n=100\\ -Participants recurited using Prolific\\ -Age range from under 19 to 69\\ -33 male, 63 female, 2 other, 2 prefer \\ not to say\\ -Mostly inexperienced with robots, except \\ domestic robots (25\%)\\ -Diverse locations and educational levels\end{tabular}                                                                                                                                               & \begin{tabular}[c]{@{}l@{}}-Slower and smooth movements with \\ conventional rotations were perceived more \\ human-like\\ -Higher human-likeness correlates with \\ preference\end{tabular}                                                                                                                                                                                                                                                                                & \begin{tabular}[c]{@{}l@{}}-No relevant effect of any demographic \\ variables\end{tabular}                                                                                                                                                                                                                                                                                                                                                                                                \\ \hline
8               & \cite{Amir}                       & \begin{tabular}[c]{@{}l@{}}-Gender\\ -Prior Experience\\ -Personality\end{tabular}                                & \begin{tabular}[c]{@{}l@{}}-Gestures   \\ -ALICE\end{tabular}                                                                              & Physical Interaction         & \begin{tabular}[c]{@{}l@{}}-Gestures such as head bowing, \\ head shaking, neck rotation and arm \\ raising were manipulated by using \\ different velocity and acceleration \\ characteristics\end{tabular}                                                                                                                  & \begin{tabular}[c]{@{}l@{}}-n=60\\ -University students and staff\\ -Age range from 20 to 57 years, M = 29.6 \\ and SD = 9.4\\ -36 male, 24 female\\ -40\% had prior interactions with robots, \\ 60\% have not interacted with robots before\end{tabular}                                                                                                                                                                                      & \begin{tabular}[c]{@{}l@{}}-The robot behavior that combines speech, \\ facial expressions, and gestures attained a \\ higher level of expressivity than the less \\ affective robot behaviors\\ -The perception of the extraverted participants \\ for the robot behavior was higher than that of \\ the introverted participants\end{tabular}                                                                                                                             & \begin{tabular}[c]{@{}l@{}}-Both gender groups have positively \\ perceived the affective expressivity of the \\ generated robot behavior\\ -An opposite-sex attraction effect of \\ human users to robots was found\end{tabular}                                                                                                                                                                                                                                                          \\ \hline
9               & \cite{Fujii}                      & -Age                                                                                                              & \begin{tabular}[c]{@{}l@{}}-Smoothness of \\ gestures\\ -NAO\end{tabular}                                                                  & Physical Interaction         & -Smoothness not described                                                                                                                                                                                                                                                                                                     & \begin{tabular}[c]{@{}l@{}}-n1=12\\ -n2=37\\ -Experiment 1: 2 male, 10 female\\ -Experiment 2: 24 participants age 15 and \\ older, thereof 11 male and 13 female; \\ 13 children under 15, thereof 10 male \\ and 3 female\end{tabular}                                                                                                                                                                                                        & \begin{tabular}[c]{@{}l@{}}-Several references on how to improve \\ interactions with the robot\end{tabular}                                                                                                                                                                                                                                                                                                                                                                & \begin{tabular}[c]{@{}l@{}}-Positive correlation between the \\ smoothness of gestures and familiarity \\ with the robot for children\\ -The smoothness of the robot’s movements \\ was positively correlated with \\ embarrassment for adults\end{tabular}                                                                                                                                                                                                                                \\ \hline
10              & \cite{Brandl}                     & \begin{tabular}[c]{@{}l@{}}-Age\\ -Gender\end{tabular}                                                            & \begin{tabular}[c]{@{}l@{}}-Speed\\ -Speed profiles\\ -Care-O-bot 3\end{tabular}                                                           & Physical Interaction         & \begin{tabular}[c]{@{}l@{}}-Speed between 0.25 and 0.75 m/s\\ -Speed profiles: constant, uniformly \\ decelerated and gradually decelerated\end{tabular}                                                                                                                                                                      & \begin{tabular}[c]{@{}l@{}}-n=30\\ -Age range from 20 to 75, AM=43.33, \\ SD = 19.01\\ -13 men, 17 women\\ -Average affinity for technology\end{tabular}                                                                                                                                                                                                                                                                                        & \begin{tabular}[c]{@{}l@{}}-Approach number had influences accepted \\ distances, increasing number of approaches \\ decreases acceptable distance\\ -No effect of robot appearance\\ -Tendency to accept longer distances if \\ participants are lying rather than standing\\ -The higher robot speed, the higher the \\ accepted distance\end{tabular}                                                                                                                    & -No effects of age and gender                                                                                                                                                                                                                                                                                                                                                                                                                                                              \\ \hline
\end{tabular}%
}
\end{sidewaystable}

\begin{figure}[ht]
    \centering
    \includegraphics[width=\textwidth]{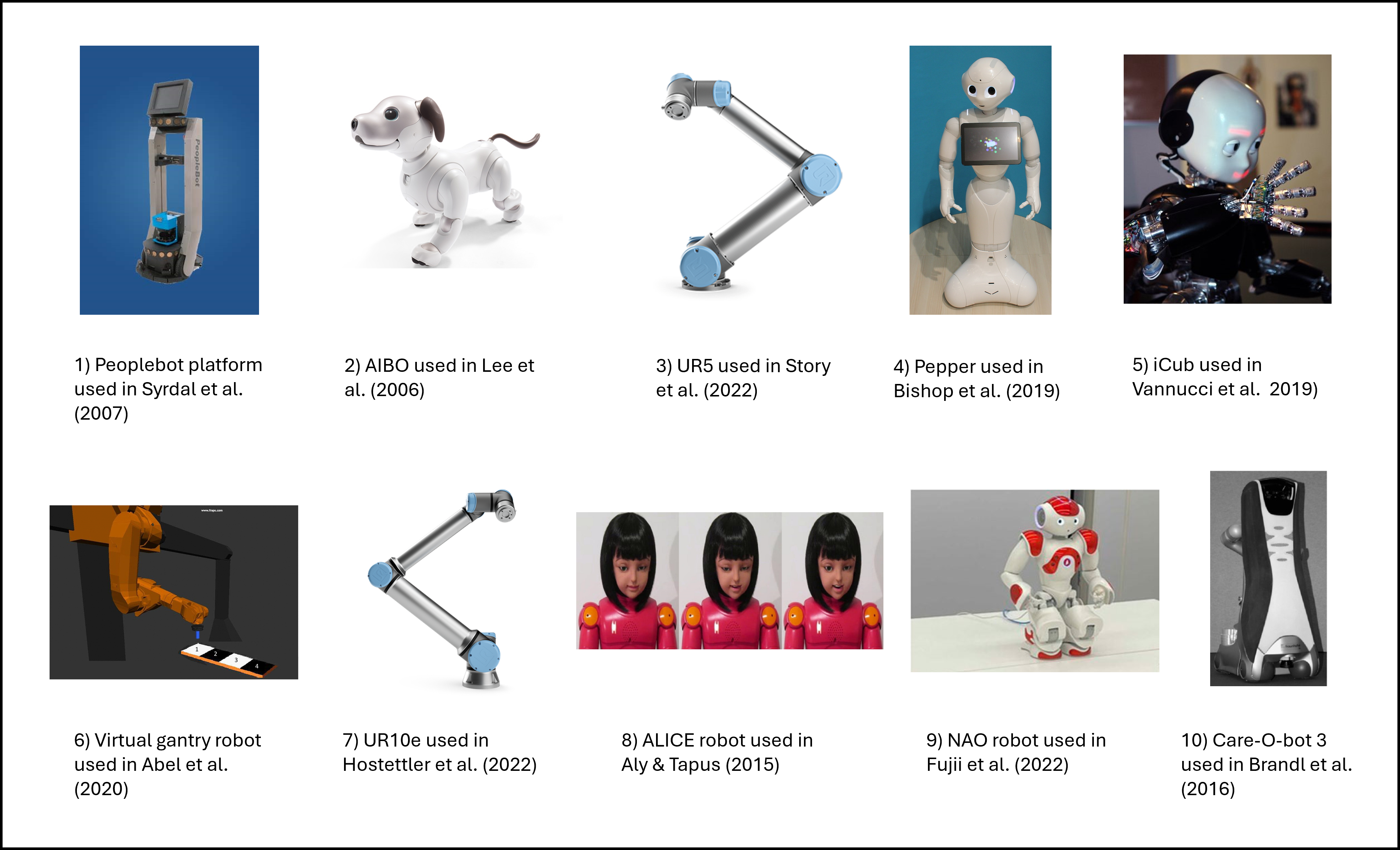}
    \caption{Robot types used in the reviewed publications.}
    \label{IllustrationRobottypes}
\end{figure}

Table~\ref{LiteratureReview} summarizes the findings of our literature review, while Figure~\ref{IllustrationRobottypes}\footnote{Sources: 1) \url{https://telepresencerobots.com/robots/adept-mobilerobots-peoplebot/}, 2)
\url{https://us.aibo.com/feature/feature1.html}, 3)
\url{https://store.clearpathrobotics.com /products/universal-robots-ur5}, 4)
\url{https://de.wikipedia.org/wiki/Pepper_\%28Roboter\%29}, 5)
\url{https://de.wikipedia.org/wiki/ICub}, 6) 
\url{https://telepresencerobots.com/robots/adept-mobilerobots-peoplebot/}, 7)
\url{https://www.universal-robots.com/products/ur10-robot/}, 8)
\url{10.1109/IROS.2015.7353789}, 9)
\url{10.3389/frobt.2022.933001}, 10)
\url{10.1002/hfm.20675}
} shows the different robots that were considered in the literature. ~\cite{Syrdal} investigated spatial preferences considering demographic data. Using a \emph{Peoplebot}, they found no effect of age, academic background, and computer proficiency on approach direction preferences, but significant gender differences as well as habituation effects after repeated interactions, indicating relevance of familiarity with the robot. Modulating motion angles, motion speed, and frequency of autonomous movements of an \emph{AIBO} robotic dog, \cite{Leearticle} found no effect of gender, but a complementarity attraction effect. With regard to speed and proxemics using an industrial UR5 robot, \cite{Story2022} found no effect of the amount of experience with industrial robots on workload or trust in a human-robot collaboration (HRC) task. \cite{Bishop2019SocialAcceptance} studied the effect of expressed robot emotion on human acceptance with a \emph{Pepper} semi-humanoid robot, considering age, gender, education, robot familiarity, and participant mood. Robot emotion was implemented by manipulating head position, voice pitch, and speaking rate, and modification of the extent of movements (``extreme'' vs. small). Regarding individual differences, while gender and education were found not to correlate with acceptance, age and mood were associated with acceptance. \cite{Vannucci} investigated cultural differences in speed adaptation during an HRI task with a \emph{iCub} humanoid robot and found no differences between samples in Italy and Japan, even though the general perception of robots between inhabitants of these countries differ substantially. \cite{Abel} compared human-like and robotic movements with a virtual gantry robot and found gender differences regarding the perception of anthropomorphic movements. \cite{Hostettler} manipulated an industrial \emph{UR10} robot's speed, movement range, smoothness, and rotations, and found effects on perceived human-likeness and human preferences, while no relevant difference was found with respect to demographic characteristics. Regarding emotion expression, a humanoid \emph{ALICE} robot was shown to support its expressivity levels with non-verbal communication cues and found opposite-sex preferences \cite{Amir}, whereas \cite{Fujii} found age differences by manipulating gesture smoothness of a semi-humanoid \emph{NAO} robot. While they found positive correlations between gesture smoothness and familiarity with the robot for children, smoothness of the robot's movements was correlated with embarrassment among adults. Lastly, \cite{Brandl} investigated approach scenarios with differing speeds and speed profiles of an ``interactive butler'' \emph{Care-O-bot 3}, and found effects of speed manipulations on accepted distances between participants and the approaching robot, but no effects of age and gender.

\section{Generation of Hypotheses}

Even though some similarities between the reviewed studies seem to exist, they largely differ in several ways. First, the robots used range from humanoid robots to articulated robot arms. Second, the type of interaction and the experimental designs correspond to varying levels of involvement of participants, and might explain some of the diverging results. Third, manipulations representing the independent variables (IV) as well as human perceptions and evaluations representing the dependent variables (DV) are not designed and implemented in standardized ways. For example, definitions of fast and slow speed differ~\cite{Leearticle, Vannucci, Hostettler, Brandl}, some studies combine movements with verbal or social cues~\cite{Bishop2019SocialAcceptance, Amir, Fujii}. DVs include variables such as perceived anthropomorphism, familiarity or general preferences, thereby adding additional complexity to the comparison of the results. And last, samples not only differ in size, but also regarding their familiarity with robots and technology in general. Some studies recruited students who are naturally younger and might have a higher affinity for current technology than older persons, whereas other studies used more balanced samples with regards to age. Still, they might not be consistent with actual robot operators' perceptions.

To build hypotheses based on the presented findings, some additional connections need to be made. First, following the findings in~\cite{Hostettler}, we use those movement types (MovTypes) that they found to be relevant and practicable with an articulated robot's degrees of freedom (DoF): speed, smoothness, and rotations. Regarding smoothness of movements, we put the implementation described in ~\cite{Hostettler} on a level with speed profiles investigated in~\cite{Brandl}. We also imply the relationships between the investigated MovTypes and perceived human-likeness presented in~\cite{Hostettler} to use the findings from~\cite{Abel} on perceived anthropomorphism, namely:

\begin{itemize}
    \item \textbf{Speed}: Slow movements are perceived more human-like than fast movements.
    \item \textbf{Smoothness}: Calm movements are perceived more human-like than nervous movements.
    \item \textbf{Rotations}: Conventional rotations are perceived more human-like than unconventional rotations.
\end{itemize}

Second, with regard to physical distances, we include the relation between gender and movement range found in our review. As general findings on personal space and gender after which women prefer larger distances to strangers~\cite{sorokowska} contradict with the opposite-sex effect found in~\cite{Amir}, we assume preference of women for a larger movement range (which can be perceived as closer approximation) . To ensure consideration of all the relations found in existing literature, we consequently add \emph{movement range} to the relevant MovTypes. Considering findings presented in \cite{Syrdal} and to include as many of the relevant relations between movements and human perceptions as possible we also investigate the robot's \emph{approach direction}, even though they have not been found to affect perceived human-likeness in~\cite{Hostettler}.

Building on the findings presented and the additional relations described above, we hence include in our hypothesis all indications found, both with regard to effects of demographic characteristics as well as MovTypes that have been shown to be relevant for differing perceptions. We thus consider \emph{five MoveTypes} (approach direction, movement range, rotations, smoothness, speed) and \emph{six demographic characteristics} (gender, location as an indication of culture, technology affinity, age, prior experience with robots, and education). Table~\ref{Overview} summarizes the hypothetical relations between robot movements and demographic characteristics used for the following explanations, and thereby represents an objective base we can use to generate our hypotheses. Accordingly, we transform the findings presented above into 30 hypotheses, and summarize these in Table~\ref{hypotheses}.

\begin{table}[ht]
\caption{Hypothetical relations between robot movements and demographic characteristics found in the literature review.}
\label{Overview}
\centering
\resizebox{\textwidth}{!}{%
\begin{tabular}{llllll}
                                                                                     & \textbf{Aproach Direction}                                                                                                                                                                           & \textbf{Movement Range}                                                                                                                                                                                                                                                              & \textbf{Rotation}                                                                                                                                                                                    & \textbf{Smoothness}                                                                                                                                                                                                                                                                                                                                                                                 & \textbf{Speed}                                                                                                                                                                                                                                                          \\ \hline
\textbf{Age}                                                                         & \cite{Hostettler}: No effect                                                                                                                                                                                         & \begin{tabular}[c]{@{}l@{}}\cite{Syrdal} \& \cite{Hostettler}: No effect\\  \cite{Bishop2019SocialAcceptance}: Negative correlations \\ between age and perceived \\ enjoyment and social \\ influence, but no direct \\ relationship between age \\ and movement range\end{tabular}                                                               & \cite{Hostettler}: No effect                                                                                                                                                                                         & \begin{tabular}[c]{@{}l@{}}\cite{Bishop2019SocialAcceptance}: Negative correlations \\ between age and perceived \\ enjoyment and social \\ influence, but no direct \\ relationship between age \\ and speed\\ \cite{Hostettler} \& \cite{Brandl}: No effect\\ \cite{Fujii}: Negative correlation \\ between age and smoothness\end{tabular}                                                                                                                               & \cite{Hostettler} \& \cite{Brandl}: No effect                                                                                                                                                                                                                                                        \\ \hline
\textbf{Gender}                                                                      & \begin{tabular}[c]{@{}l@{}}\cite{Syrdal}: Males allow the \\ robot to approach closer \\ when the robot approaches \\ from the side, women \\ when the robot approaches \\ directly from the front\end{tabular} & \begin{tabular}[c]{@{}l@{}}\cite{Leearticle}, \cite{Bishop2019SocialAcceptance} \& \cite{Hostettler}: No effect\\ \cite{Amir}: Opposite-attraction effect, \\ therefore preference of large \\ movement range for females \\ and vice versa\end{tabular}                                                                                                         & \begin{tabular}[c]{@{}l@{}}\cite{Abel}: Females attributed more \\ anthropomorphic features to \\ robotic movements, therefore \\ potentially less sensitive to \\ robot rotations\\ \cite{Hostettler}: No effect\end{tabular} & \begin{tabular}[c]{@{}l@{}}\cite{Bishop2019SocialAcceptance}, \cite{Hostettler} \& \cite{Brandl}: No effect\\ \cite{Abel}: Males were sensitive to \\ differences regarding human-\\ like vs. robotic movements, \\ whereas females were not\\ \cite{Abel}: Females attributed more \\ anthropomorphic features to \\ robotic movements, therefore \\ potentially less sensitive to \\ robot smoothness\\ \cite{Amir}: Opposite-attraction effect, \\ unclear effect of smoothness\end{tabular} & \begin{tabular}[c]{@{}l@{}}\cite{Leearticle}, \cite{Hostettler} \& \cite{Brandl}: No effect\\ \cite{Abel}: Females attributed more \\ anthropomorphic features to \\ robotic movements, therefore \\ potentially less sensitive to \\ robot speed\\ \cite{Amir}: Opposite-attraction effect, \\ unclear effect of speed\end{tabular}      \\ \hline
\textbf{\begin{tabular}[c]{@{}l@{}}Prior Experience \\ with robots\end{tabular}}     & \cite{Hostettler}: No effect                                                                                                                                                                                        & \begin{tabular}[c]{@{}l@{}}\cite{Story2022} \& \cite{Hostettler}: No effect\\  \cite{Bishop2019SocialAcceptance}: Participants more familiar \\ with robots felt they had less \\ sufficient knowledge to use \\ the system and found the \\ robot less safe, but no direct \\ relationship between \\ familiarity and movement \\ range\end{tabular} & \cite{Hostettler}: No effect                                                                                                                                                                                         & \begin{tabular}[c]{@{}l@{}}\cite{Bishop2019SocialAcceptance}: Participants more familiar \\ with robots felt they had less \\ sufficient knowledge to use \\ the system and found the \\ robot less safe, but no direct \\ relationship between \\ familiarity and smoothness\\ \cite{Hostettler}: No effect\end{tabular}                                                                                                                           & \begin{tabular}[c]{@{}l@{}}\cite{Story2022} \& \cite{Hostettler}: No effect\\ \cite{Bishop2019SocialAcceptance}: Participants more familiar \\ with robots felt they had less \\ sufficient knowledge to use \\ the system and found the \\ robot less safe, but no \\ direct relationship between \\ familiarity and speed\end{tabular} \\ \hline
\textbf{\begin{tabular}[c]{@{}l@{}}Computer /   \\ Technology affinity\end{tabular}} & n/a                                                                                                                                                                                                  & \cite{Syrdal}: No effect                                                                                                                                                                                                                                                                         & n/a                                                                                                                                                                                                  & n/a                                                                                                                                                                                                                                                                                                                                                                                                 & n/a                                                                                                                                                                                                                                                                     \\ \hline
\textbf{Location /   Culture}                                                        & \cite{Hostettler}: No effect                                                                                                                                                                                         & \cite{Vannucci} \& \cite{Hostettler}: No effect                                                                                                                                                                                                                                                                      & \cite{Hostettler}: No effect                                                                                                                                                                                         & \cite{Vannucci} \& \cite{Hostettler}: No effect                                                                                                                                                                                                                                                                                                                                                                                     & \cite{Vannucci} \& \cite{Hostettler}: No effect                                                                                                                                                                                                                                                         \\ \hline
\textbf{Education}                                                                   & \cite{Hostettler}: No effect                                                                                                                                                                                         & \cite{Syrdal}, \cite{Bishop2019SocialAcceptance} \& \cite{Hostettler}: No effect                                                                                                                                                                                                                                                                   & \cite{Hostettler}: No effect                                                                                                                                                                                         & \cite{Bishop2019SocialAcceptance} \& \cite{Hostettler}: No effect                                                                                                                                                                                                                                                                                                                                                                                     & \cite{Hostettler}: No effect                                                                                                                                                                                                                                                            \\ \hline
                                                                                     &                                                                                                                                                                                                      &                                                                                                                                                                                                                                                                                      &                                                                                                                                                                                                      &                                                                                                                                                                                                                                                                                                                                                                                                     &                                                                                                                                                                                                                                                                        
\end{tabular}%
}
\end{table}

\begin{table}[ht]
\caption{Generated hypotheses for all relations between movements and demographic characteristics.}
\label{hypotheses}
\centering
\resizebox{\textwidth}{!}{%
\begin{tabular}{llllllll}
                                                                                     & \textbf{Aproach Direction}                                                                                         & \textbf{Movement Range}                                                                                               & \textbf{Rotation}                                                                                         & \textbf{Smoothness}                                                                                            & \textbf{Speed}                                                                                            &  &  \\ \cline{1-6}
\textbf{Age}                                                                         & No effect expected                                                                                                 & No effect expected                                                                                                    & No effect expected                                                                                        & \begin{tabular}[c]{@{}l@{}}Younger people \\ prefer smooth \\ movements\end{tabular}                           & No effect expected                                                                                        &  &  \\ \cline{1-6}
\textbf{Gender}                                                                      & \begin{tabular}[c]{@{}l@{}}Males prefer approach \\ from the side\\ Females prefer direct \\ approach\end{tabular} & \begin{tabular}[c]{@{}l@{}}Males prefer small \\ movement range\\ Females prefer large \\ movement range\end{tabular} & \begin{tabular}[c]{@{}l@{}}Males prefer less \\ rotations\\ Females prefer more \\ rotations\end{tabular} & \begin{tabular}[c]{@{}l@{}}Males prefer smooth \\ movements\\ Females prefer nervous \\ movements\end{tabular} & \begin{tabular}[c]{@{}l@{}}Males prefer slow \\ movements\\ Females prefer fast \\ movements\end{tabular} &  &  \\ \cline{1-6}
\textbf{\begin{tabular}[c]{@{}l@{}}Prior Experience \\ with robots\end{tabular}}     & No effect expected                                                                                                 & \begin{tabular}[c]{@{}l@{}}Experienced people \\ prefer small \\ movement range\end{tabular}                          & No effect expected                                                                                        & \begin{tabular}[c]{@{}l@{}}Experienced people \\ prefer smooth \\ movements\end{tabular}                       & \begin{tabular}[c]{@{}l@{}}Experienced people \\ prefer slow \\ movements\end{tabular}                    &  &  \\ \cline{1-6}
\textbf{\begin{tabular}[c]{@{}l@{}}Computer /   \\ Technology affinity\end{tabular}} & \begin{tabular}[c]{@{}l@{}}Not yet investigated \\ No effect expected\end{tabular}                                 & No effect expected                                                                                                    & \begin{tabular}[c]{@{}l@{}}Not yet investigated \\ No effect expected\end{tabular}                        & \begin{tabular}[c]{@{}l@{}}Not yet investigated \\ No effect expected\end{tabular}                             & \begin{tabular}[c]{@{}l@{}}Not yet investigated \\ No effect expected\end{tabular}                        &  &  \\ \cline{1-6}
\textbf{Location /   Culture}                                                        & No effect expected                                                                                                 & No effect expected                                                                                                    & No effect expected                                                                                        & No effect expected                                                                                             & No effect expected                                                                                        &  &  \\ \cline{1-6}
\textbf{Education}                                                                   & No effect expected                                                                                                 & No effect expected                                                                                                    & No effect expected                                                                                        & No effect expected                                                                                             & No effect expected                                                                                        &  &  \\ \cline{1-6}
                                                                                     &                                                                                                                    &                                                                                                                       &                                                                                                           &                                                                                                                &                                                                                                           &  & 
\end{tabular}%
}
\end{table}

To further examine the relationships between robot movements and individual differences, we build our experimental design upon these findings and present our methodological approach in Section~\ref{Methods}.

\section{Methods} \label{Methods}

We define the type of interaction between robots and people according to~\cite{Story2022}: ``HRI considers any situation where a robot’s actions, or inactions, result in a reaction from a human (and vice versa)''. Accordingly, we focus on HRI and do not make a claim about effects in HRC settings which ``consider any situation where the robot and the human work together to complete a task.'' Our experiment is Web-based with the objective of reaching a large sample size and therefore does not allow to let participants actually work together with the robot. However, HRI, defined as above includes a wide spectrum where, for instance, on industrial shopfloors operators work in proximity to and observe, perceive, and evaluate robots.

To capture the effect of an industrial robot arm's movements on individual perception, we replicate to a certain extent the study presented in~\cite{Hostettler}, as their approach allows to isolate the effects of distinct robot movements on human perception. While their main goal was to measure the perceived human-likeness of an articulated robot's movements, they also calculated correlations to the participants' expressed preferences. Also, they have published the videos used in their experiments showing a UR 10e robot moving in two conditions per MovType, matching most of the movements laid out in our review in Section~\ref{sec2}. We therefore combine their videos of robot movements, a question regarding participants' preferences, and additional queries of demographic criteria.

\subsection{Experimental Procedure}\label{ExperimentalProcedure}

Using the MovTypes as well as the corresponding videos published in~\cite{Hostettler}, we conducted a Web-based experiment. Participants were asked to state their preferences for five different movement behaviours, with two conditions for each movement:

\begin{itemize}
    \item Speed: slow (A) vs. fast (B) movement
    \item Smoothness: smooth (A) vs. nervous (B) movement
    \item Rotations: conventional (A) vs. unconventional (B) rotation
    \item Movement range: small (A) vs. large (B) movement range
    \item Approach direction: from the side (A) vs. straight towards the participant (B)
\end{itemize}

\begin{figure}[ht]
    \centering
    \includegraphics[width=\linewidth]{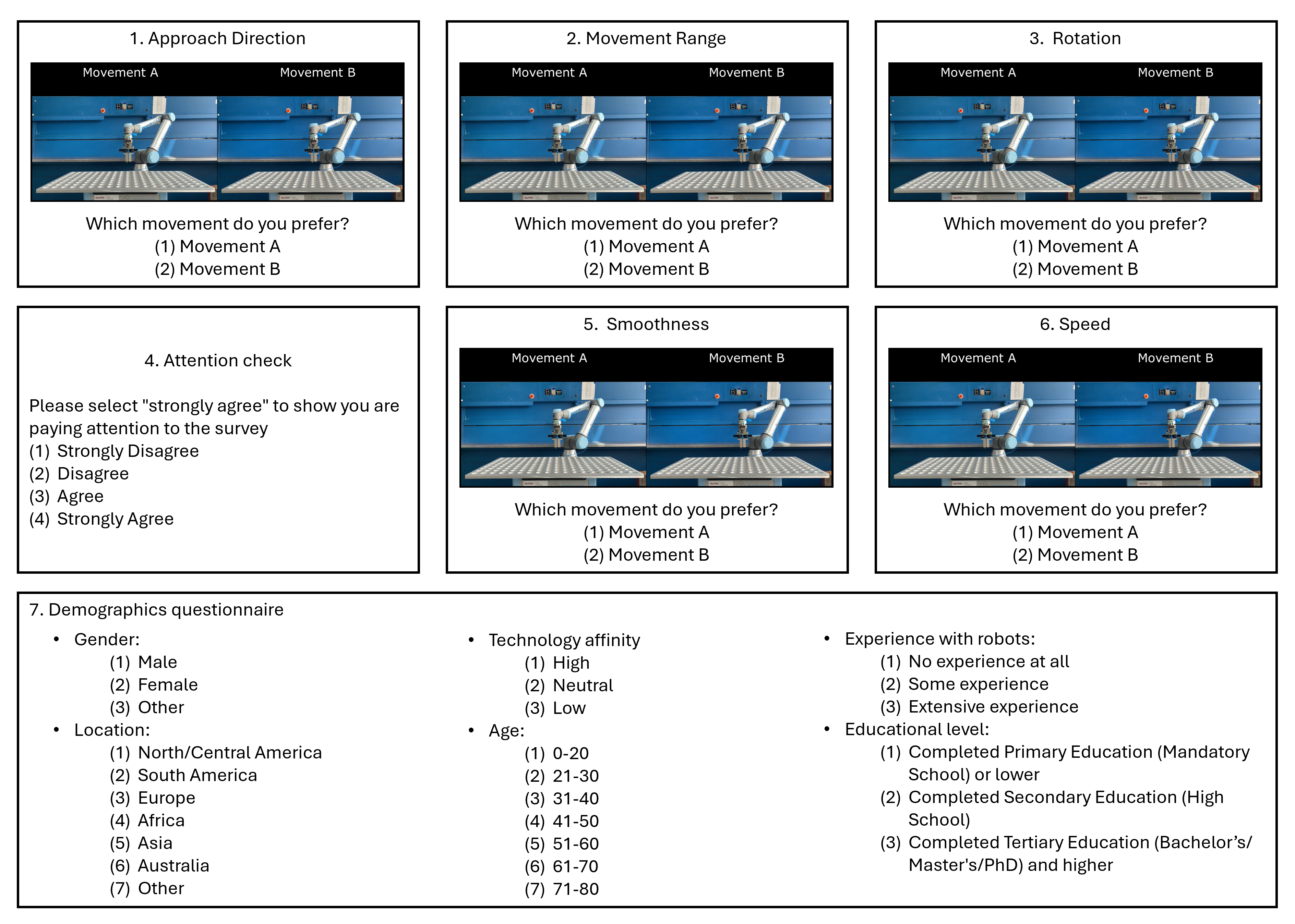}
    \caption{Study structure and questions used.}
    \label{studystructure}
\end{figure}

The videos were used unmodified from~\cite{Hostettler}, and we hence refer to their explanations regarding the detailed movement design. Participants were recruited through \emph{Prolific Academia} and asked to complete the study as illustrated in Figure~\ref{studystructure}.

Designed as within-subject experiment, each participant rated all five videos and saw both conditions per video---movement A on the left side and movement B on the right side---playing consecutively. To prevent selection biases, we have randomized the order of the videos (MovTypes) as well as the order of the conditions (A-B / B-A) within each video. Participants were then asked to state their preference for either movement A or movement B (question: ``Which movement behavior do you prefer?''). After rating all videos, participants were asked to disclose their gender, location, technology affinity, age, experience with robots and educational level. To ensure participants' attentiveness, an attention check was included after the third video. The study was conducted in February 2024 with a mean duration of 5 minutes and 50 seconds.

\subsection{Participants}\label{participants}

To determine the sample size required to detect meaningful effects, we have conducted a priori power analyses (\emph{G*Power 3.1}) assuming intermediate effect sizes with repeated measures. Participants were recruited through Prolific Academia, and pre-screened for English being the participants' first language and an approximative uniform distribution of gender. 

1’027 Participants were recruited, whereof 97 participants were not included in the final analysis due to timeouts or a failed attention check. In addition, some groups were underrepresented in our sample, and to ensure sufficient sample sizes per group we not included participants in the following cases:

\begin{itemize}
    \item We have ignored timeouts and if participants failed the attention check (n=36)
    \item Gender: We have ignored all participants who identified as ``other'' (n=9)
    \item Location: We have ignored participants form South America (n=19), Asia (n=4), Australia (n=7) and ``other'' (n=22)
\end{itemize}

Moreover, some groups of demographic characteristics were underrepresented. To ensure analyzability of all groups included, we have grouped these categories as follows:

\begin{itemize}
    \item Technology affinity: We have grouped participants with low (n=42) and neutral (n=431) technology affinity in the group ``lower technology affinity'', compared to the group ``high technology affinity'' (n=457).
    \item Age: We have grouped participants below 20 years old (n=34) and 21-30 years old (n=506) in the group ``30 years old and below'', and the groups from 31-80 years old to the group ``31 years old and above'' (n=390).
    \item Experience with robots: We have grouped participants with some experience (n=521) and extensive experience (n=41) in the group ``higher level of experience'', compared to the group with ``no experience at all'' (n=368).
    \item Education: We have grouped participants with completed primary education or lower (n=8) and participants with completed secondary education (n=283) to the group ``lower educational level'', compared to the group ``higher educational level'' with completed tertiary education and higher (n=639).
\end{itemize}

Using these constraints, a final sample size of n=930 resulted. This final sample included an almost equal distribution of gender (49.6\% identified as male (n=461), 50.4\% as female (n=469)), and a range of locations with a majority from Europe (69.2\%, n=644), 19.4\% from Africa (n=182) and 11.1\% from North America (n=104). Participants' technology affinity was almost equally distributed (50.9\% with a lower level of technology affinity, 49.1\% with a higher level of technology affinity), and a mix of ages (58.1\% in the age category 30 years old and below , 41.9\% in age category 31 years old and above). Most participants had a higher level of experience with robots (60.4\%) and education (68.7\% with completed tertiary education or higher).

\subsection{Data Analysis}\label{Analysis}

The resulting data was analysed using the \emph{IBM SPSS} software. To assess the impact of several predictors on a dichotomous dependent variable, a Generalized Estimating Equations (GEE) binary logistic regression was conducted where the rating corresponds to the participants’ preference for one of the two movement conditions. The independent variables in the model included MovTypes (1-5), gender (male; female), location (North America; Europe; Africa), technology affinity (lower, higher), age ($\le$30; $\ge$31), experience with robots (lower; higher), and education (lower, higher). The analysis utilized a within-subjects design to account for repeated measurements, with the first preference/rating being the reference category. The working correlation structure was unspecified, using the model's default selection. To evaluate the models’ fits, we used the Hosmer-Lemeshow test. All models show no significant effects in the Hosmer-Lemeshow test (all p values $>$ 0.1), except for the model of the individual video Smoothness (${\chi}^2(8) = 20.20, p = .010$) and Speed (${\chi}^2(8) = 20.37, p = .009$). These significant effects can potentially be explained by the large sample size of this study, but the results of these analyses should be interpreted with care. Note, however, that both models result in strong R2 values indicating the models’ ability to explain the variance in the data.

\section{Results}
\label{Results}

\begin{figure}[ht]
    \centering
    \includegraphics[height=4.7cm]{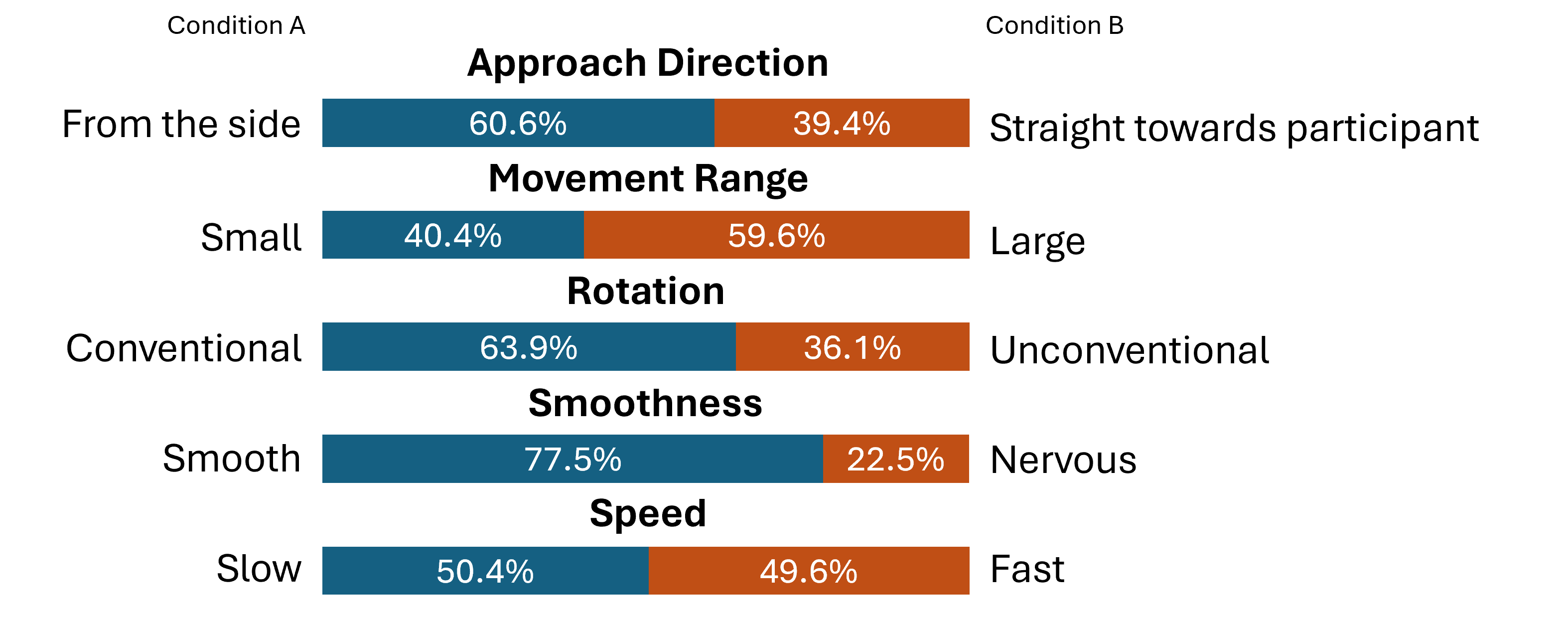}
    \caption{Preference ranking per MovType (n=930).}
    \label{Haeufigkeiten}
\end{figure}

The fundamentals of our analysis are participants' preference ratings for either condition A or B per MovType. Analysing the entire sample's preferences, Figure~\ref{Haeufigkeiten} shows the preferred conditions per MovType, irrespective of group-specific differences: The results demonstrate an overall preference for the approach from the side, a large movement range, conventional rotations, and smooth movements. Regarding speed, preferences for slow and fast movements were almost equally distributed.

Across all MovTypes, the GEE binary logistic regression (with p level $<$ 0.05) reveals that location, (Wald ${\chi}^2(2) = 15.29, p < .0011$), gender, (Wald ${\chi}^2(1) = 16.23, p < .001$), and experience (Wald ${\chi}^2(1) = 6.72, p < .01$) are significant predictors, as shown in Table~\ref{ResultsTable1}. This indicates that changes in these predictors are associated with different probabilities of selecting the reference rating category. In more detail, compared to participants from \emph{Africa}, participants from \emph{North America} ($p = .043$) and \emph{Europe} ($p < .001$) are 21\% and 28.7\% less likely to prefer condition B (straight approach, large movement range, unconventional rotations, nervous, fast movements) relative to condition A (see Table \ref{ResultsTable2}). Concerning gender, \emph{male} participants ($p < .001$) are 31.2\% more likely to prefer condition B relative to condition A, relative to \emph{female} participants. Finally, people with more experience with robots ($p = .009$) are 16\% less likely to select condition B relative to condition A, relative to individuals with lower levels of experience (see Table~\ref{ResultsTable2}). Age, education, and technology affinity are not significant predictors, suggesting that within the context of this model, they do not influence the likelihood of selecting the reference rating category.

\begin{table}[ht]
\label{ResultsTable1}
\caption{Tests of Model’s Effects.}
\begin{tabular}{llll}
\hline
\multirow{2}{*}{Source} & \multicolumn{3}{l}{Type   III} \\ \cline{2-4} 
                        & Wald-Chi-Square & df & Sig.    \\ \hline
(Constant)              & 52.918          & 1  & $<$.001 \\ \hline
MovType                 & 299.999         & 4  & .000    \\ \hline
Age                     & 2.266           & 1  & .132    \\ \hline
Experience              & 6.727           & 1  & .009    \\ \hline
Gender                  & 16.231          & 1  & $<$.001 \\ \hline
Location                & 15.298          & 2  & $<$.001 \\ \hline
Education               & .021            & 1  & .885    \\ \hline
TechAffinity            & .412            & 1  & .521    \\ \hline
\end{tabular}%

\end{table}

\begin{table}[ht]
\label{ResultsTable2}
\caption{Parameter estimates from the GEE binary logistic regression model.}
\resizebox{\textwidth}{!}{%
\begin{tabular}{lllllll}
\hline
\multirow{2}{*}{Parameter}   & \multirow{2}{*}{\begin{tabular}[c]{@{}l@{}}Regression \\ Coefficient B\end{tabular}} & \multirow{2}{*}{Std.-Error} & \multicolumn{3}{l}{Hypothesis Testing}  & \multirow{2}{*}{Exp(B)} \\ \cline{4-6}
                             &                                                                                      &                             & Wald-Chi-Squared & df & Sig.            &                         \\ \hline
(Constant)                   & .101                                                                                 & .1213                       & .698             & 1  & .403            & 1.107                   \\ \hline
{[}MovType=ApproachDir{]}    & -.419                                                                                & .0924                       & 20.619           & 1  & \textless{}.001 & .657                    \\ \hline
{[}MovType=MovRange{]}       & .409                                                                                 & .0947                       & 18.633           & 1  & \textless{}.001 & 1.505                   \\ \hline
{[}MovType=Rotation{]}       & -.558                                                                                & .0975                       & 32.779           & 1  & \textless{}.001 & .572                    \\ \hline
{[}MovType=Smooth{]}         & -1.232                                                                               & .0944                       & 170.443          & 1  & .000            & .292                    \\ \hline
{[}MovType=Speed{]}          & 0a                                                                                   & .                           & .                & .  & .               & 1                       \\ \hline
{[}Age(below31){]}           & .099                                                                                 & .0658                       & 2.266            & 1  & .132            & 1.104                   \\ \hline
{[}Age(above31){]}           & 0a                                                                                   & .                           & .                & .  & .               & 1                       \\ \hline
{[}Experience(low){]}        & -.174                                                                                & .0672                       & 6.727            & 1  & .009            & .840                    \\ \hline
{[}Experience(high){]}       & 0a                                                                                   & .                           & .                & .  & .               & 1                       \\ \hline
{[}Gender(male){]}           & .272                                                                                 & .0674                       & 16.231           & 1  & \textless{}.001 & 1.312                   \\ \hline
{[}Gender(female){]}         & 0a                                                                                   & .                           & .                & .  & .               & 1                       \\ \hline
{[}Location(NorthAmerica){]} & -.236                                                                                & .1167                       & 4.094            & 1  & .043            & .790                    \\ \hline
{[}Location(Europe){]}       & -.339                                                                                & .0868                       & 15.240           & 1  & \textless{}.001 & .713                    \\ \hline
{[}Location(Africa){]}       & 0a                                                                                   & .                           & .                & .  & .               & 1                       \\ \hline
{[}Education(low){]}         & -.010                                                                                & .0691                       & .021             & 1  & .885            & .990                    \\ \hline
{[}Education(high){]}        & 0a                                                                                   & .                           & .                & .  & .               & 1                       \\ \hline
{[}TechAffinity(low){]}      & .044                                                                                 & .0683                       & .412             & 1  & .521            & 1.045                   \\ \hline
{[}TechAffinity(high){]}     & 0a                                                                                   & .                           & .                & .  & .               & 1                       \\ \hline
(Scale)                      & 1                                                                                    &                             &                  &    &                 &                         \\ \hline
\end{tabular}%
}
\end{table}

In addition, MovType emerges as a significant predictor (Wald ${\chi}^2$(4) = 299.99, p $<$ .001), indicating that the type of movement significantly influences the preference for either condition A or B. The specific nature of this influence was further explored through pairwise comparisons to discern the differences in the likelihood of selecting the preference's reference category among the various MovTypes, and reveals significant differences between the majority of different MovTypes. These findings suggest that certain MovTypes are statistically more or less likely to result in participants expressing a higher preference for the reference category compared to others (see Table~\ref{ResultsTable3}).

\begin{table}[ht]
\label{ResultsTable3}
\caption{Pairwise comparisons.}
\resizebox{\textwidth}{!}{%
\begin{tabular}{llllllll}
\hline
\multirow{2}{*}{MovType(I)} & \multirow{2}{*}{MovType(J)} & \multirow{2}{*}{\begin{tabular}[c]{@{}l@{}}MeanDifference \\ (I-J)\end{tabular}} & \multirow{2}{*}{Std.-Error} & \multirow{2}{*}{df} & \multirow{2}{*}{Sig.} & \multicolumn{2}{l}{\begin{tabular}[c]{@{}l@{}}95\% Wald-Confidence-\\ Interval for the Difference\end{tabular}} \\ \cline{7-8} 
                            &                             &                                                                                  &                             &                     &                       & Lower                                                  & Upper                                                  \\ \hline
ApproachDir                 & MovRange                    & -.20                                                                             & .022                        & 1                   & .000                  & -.25                                                   & -.16                                                   \\
                            & Rotation                    & .03                                                                              & .024                        & 1                   & .174                  & -.01                                                   & .08                                                    \\
                            & Smooth                      & .17                                                                              & .021                        & 1                   & \textless{}.001       & .13                                                    & .21                                                    \\
                            & Speed                       & -.10                                                                             & .023                        & 1                   & \textless{}.001       & -.15                                                   & -.06                                                   \\ \hline
MovRange                    & ApproachDir                 & .20                                                                              & .022                        & 1                   & .000                  & .16                                                    & .25                                                    \\
                            & Rotation                    & .24                                                                              & .023                        & 1                   & .000                  & .19                                                    & .28                                                    \\
                            & Smooth                      & .38                                                                              & .022                        & 1                   & .000                  & .33                                                    & .42                                                    \\
                            & Speed                       & .10                                                                              & .023                        & 1                   & \textless{}.001       & .06                                                    & .15                                                    \\ \hline
Rotation                    & ApproachDir                 & -.03                                                                             & .024                        & 1                   & .174                  & -.08                                                   & .01                                                    \\
                            & MovRange                    & -.24                                                                             & .023                        & 1                   & .000                  & -.28                                                   & -.19                                                   \\
                            & Smooth                      & .14                                                                              & .021                        & 1                   & \textless{}.001       & .10                                                    & .18                                                    \\
                            & Speed                       & -.14                                                                             & .024                        & 1                   & \textless{}.001       & -.18                                                   & -.09                                                   \\ \hline
Smooth                      & ApproachDir                 & -.17                                                                             & .021                        & 1                   & \textless{}.001       & -.21                                                   & -.13                                                   \\
                            & MovRange                    & -.38                                                                             & .022                        & 1                   & .000                  & -.42                                                   & -.33                                                   \\
                            & Rotation                    & -.14                                                                             & .021                        & 1                   & \textless{}.001       & -.18                                                   & -.10                                                   \\
                            & Speed                       & -.28                                                                             & .020                        & 1                   & .000                  & -.32                                                   & -.24                                                   \\ \hline
Speed                       & ApproachDir                 & .10a                                                                             & .023                        & 1                   & \textless{}.001       & .06                                                    & .15                                                    \\
                            & MovRange                    & -.10                                                                             & .023                        & 1                   & \textless{}.001       & -.15                                                   & -.06                                                   \\
                            & Rotation                    & .14                                                                              & .024                        & 1                   & \textless{}.001       & .09                                                    & .18                                                    \\
                            & Smooth                      & .28                                                                              & .020                        & 1                   & .000                  & .24                                                    & .32                                                    \\ \hline
\end{tabular}%
}
\end{table}

Note, however, that the pairwise differences between various MovTypes do not allow us to answer the main research question, that is whether demographic characteristics do predict the preference (condition A vs. B) separately for each MovType. Therefore, building upon the preliminary findings from the GEE approach, which treated MovType as a singular predictor concerning the dependent variable \emph{preference rating}, we aimed to extend our understanding of the predictors' effects for each specific MovType. The logic of the initial GEE analysis is similar to a Multivariate Analysis of Variance (MANOVA), where the binary outcome of the preference is investigated across the different MovTypes. To overcome this constraint and for a more granular analysis, a series of binary logistic regressions was performed. Each MovType’s preference ratings (Approach Direction, Movement Range, Rotation, Speed, and Smoothness) were, now, \emph{individually} analysed as dependent variables. This step allows for the assessment of the influence of the IVs (age, experience, gender, location, education, and technology affinity) on the likelihood of condition A or B being rated as preferred. In these subsequent analyses, the last category of each independent variable (female, Africa, high technology affinity, 31 years old and above, higher level of experience, higher educational level) served as the reference category, in line with common practice in logistic regression to provide a benchmark for comparison. The outcome of each regression model was interpreted relative to the reference categories of the independent variables, thereby allowing for the estimation of odds ratios that reflect the likelihood of a specific condition per MovType being preferred, given a participant's demographic profile. To interpret these likelihoods correctly, we additionally present the frequencies of preference ratings per group, as higher odds ratios for a condition do not necessarily result in a majority share for the respective condition. The group-specific frequencies per MovType can be found in Table~\ref{group-specificFrequencies} in Appendix~\ref{AppendixA}.

\subsection{Approach Direction}

A binary logistic regression analysis was conducted to predict preferences in the first video (condition B vs. A) based on gender, location, experience, age, technology affinity, and education. The overall model is not statistically significant (${\chi}^2(7) = 13.55, p = .060$). The model explains between 1.4\% (Cox and Snell $R^2$) to 2.0\% (Nagelkerke $R^2$) of the variance in preference ratings and correctly classifies 60.3\% of cases. \emph{Location(Europe)} ($b = -.36, p = .047, OR = .70$) and \emph{age} ($b = .30, p = .036, OR = 1.35$) are significant predictors in the model (see Table~\ref{ResultsTable4}). Participants in Europe relative to Africa have $\sim$30\% lower odds of preferring condition B (straight approach) compared to condition A (approach from the side). Additionally, with increasing age, the odds of preferring the straight approach decreases by 35\%. In other words, European participants are significantly less likely to prefer the straight approach, compared to African participants, and contrary, participants 30 years old and below prefer the straight approach more likely, compared to participants 31 years old and above.

Taking into consideration group-specific frequencies, the direction of the effects remains unchanged. Regarding location, a slight majority of participants from Africa (53.85\%) still prefers the approach from the side.

\begin{table}[ht]
\label{ResultsTable4}
\caption{Model parameters for MovType \emph{Approach Direction}.}
\resizebox{\textwidth}{!}{%
\begin{tabular}{lllllll}
\hline
                  & \begin{tabular}[c]{@{}l@{}}Regression \\ Coefficient B\end{tabular} & \begin{tabular}[c]{@{}l@{}}Standard \\ Error\end{tabular} & Wald  & df & Sig. & Exp(B) \\ \hline
Gender(Male)      & .217                                                                & .147                                                      & 2.168 & 1  & .141 & 1.242  \\ \hline
Location          &                                                                     &                                                           & 4.235 & 2  & .120 &        \\ \hline
Location(Europe)  & -.156                                                               & .254                                                      & .380  & 1  & .538 & .855   \\ \hline
Location(Africa)  & -.355                                                               & .179                                                      & 3.940 & 1  & .047 & .701   \\ \hline
Experience(low)   & -.145                                                               & .147                                                      & .984  & 1  & .321 & .865   \\ \hline
Age(low)          & .296                                                                & .141                                                      & 4.415 & 1  & .036 & 1.345  \\ \hline
TechAffinity(low) & .005                                                                & .149                                                      & .001  & 1  & .972 & 1.005  \\ \hline
Education(low)    & -.029                                                               & .149                                                      & .039  & 1  & .844 & .971   \\ \hline
Constant          & -.391                                                               & .216                                                      & 3.265 & 1  & .071 & .676   \\ \hline
\end{tabular}%
}
\end{table}

\subsection{Movement range}
A binary logistic regression analysis was conducted to predict preferences in the second video (condition B vs. A) based on gender, location, experience, age, technology affinity, and education. The overall model is not statistically significant (${\chi}^2(7) = 10.55, p = .160$). The model explains between 1.1\% (Cox and Snell $R^2$) and 1.5\% (Nagelkerke $R^2$) of the variance in preference ratings and correctly classifies 58.8\% of cases. \emph{Experience} ($b = -.33, p = .023, OR = .72$) is a significant predictor in the model (see Table~\ref{ResultsTable5}). With increasing level of experience with robots, the odds of preferring condition B (large movement range) increases by 28\% compared to preferring condition A (small movement range). Accordingly, participants with higher levels of experience are more likely to prefer a large movement range.

Again, consideration of group-specific frequencies do not reveal any differences regarding the preferred conditions. In line with the entire sample, all groups prefer the large movement range.

\begin{table}[ht]
\label{ResultsTable5}
\caption{Model parameters for MovType \emph{Movement Range}.}
\resizebox{\textwidth}{!}{%
\begin{tabular}{lllllll}
\hline
                  & \begin{tabular}[c]{@{}l@{}}Regression \\ Coefficient B\end{tabular} & \begin{tabular}[c]{@{}l@{}}Standard \\ Error\end{tabular} & Wald   & df & Sig.    & Exp(B) \\ \hline
Gender(Male)      & .132                                                                & .146                                                      & .818   & 1  & .366    & 1.141  \\ \hline
Location          &                                                                     &                                                           & .610   & 2  & .737    &        \\ \hline
Location(Europe)  & -.148                                                               & .257                                                      & .334   & 1  & .564    & .862   \\ \hline
Location(Africa)  & -.137                                                               & .182                                                      & .573   & 1  & .449    & .872   \\ \hline
Experience(low)   & -.330                                                               & .145                                                      & 5.171  & 1  & .023    & .719   \\ \hline
Age(low)          & -.265                                                               & .140                                                      & 3.578  & 1  & .059    & .767   \\ \hline
TechAffinity(low) & .084                                                                & .148                                                      & .321   & 1  & .571    & 1.087  \\ \hline
Education(low)    & -.176                                                               & .147                                                      & 1.441  & 1  & .230    & .838   \\ \hline
Constant          & .735                                                                & .219                                                      & 11.275 & 1  & $<$.001 & 2.086  \\ \hline
\end{tabular}%
}
\end{table}

\subsection{Rotation}

A binary logistic regression analysis was conducted to predict preferences in the third video (condition B vs. A) based on gender, location, experience, age, technology affinity, and education. The overall model is not statistically significant (${\chi}^2(7) = 4.18, p = .759$). The model explains between 0.4\% (Cox and Snell $R^2$) and 0.6\% (Nagelkerke $R^2$) of the variance in preference ratings and correctly classifies 63.9\% of cases. None of the predictors in the model are statistically significant (all p values $>$ 0.05), and demographic characteristics therefore do not have an influence on the participants’ preferences for neither conventional nor unconventional rotations (see Table~\ref{ResultsTable6}). Accordingly, all group-specific frequencies correspond to the entire sample's preference for conventional rotations.

\begin{table}[ht]
\label{ResultsTable6}
\caption{Model parameters for MovType \emph{Rotation}.}
\resizebox{\textwidth}{!}{%
\begin{tabular}{lllllll}
\hline
                  & \begin{tabular}[c]{@{}l@{}}Regression \\ Coefficient B\end{tabular} & \begin{tabular}[c]{@{}l@{}}Standard \\ Error\end{tabular} & Wald  & df & Sig. & Exp(B) \\ \hline
Gender(Male)      & .008                                                                & .149                                                      & .003  & 1  & .959 & 1.008  \\ \hline
Location          &                                                                     &                                                           & 2.507 & 2  & .285 &        \\ \hline
Location(Europe)  & -.200                                                               & .258                                                      & .603  & 1  & .437 & .819   \\ \hline
Location(Africa)  & -.285                                                               & .180                                                      & 2.500 & 1  & .114 & .752   \\ \hline
Experience(low)   & -.127                                                               & .148                                                      & .737  & 1  & .391 & .880   \\ \hline
Age(low)          & -.027                                                               & .142                                                      & .037  & 1  & .847 & .973   \\ \hline
TechAffinity(low) & .085                                                                & .150                                                      & .318  & 1  & .573 & 1.088  \\ \hline
Education(low)    & .030                                                                & .150                                                      & .040  & 1  & .841 & 1.031  \\ \hline
Constant          & -.343                                                               & .218                                                      & 2.471 & 1  & .116 & .710   \\ \hline
\end{tabular}%
}
\end{table}

\subsection{Smoothness}

A binary logistic regression analysis was conducted to predict preferences in the fourth video (condition B vs. A) based on gender, location, experience, age, technology affinity, and education. The overall model is statistically significant (${\chi}^2(7) = 32.18, p < .001$). The model explains between 3.4\% (Cox and Snell $R^2$) and 5.2\% (Nagelkerke $R^2$) of the variance in preference ratings and correctly classifies 77.5\% of cases. \emph{Gender} ($b = .50, p = .004, OR = 1.66$), location \emph{North America} ($b = -.81, p = .006, OR = .44$), and location \emph{Europe} ($b = -1.01, p < .001, OR = .36$) are significant predictors in the model (see Table~\ref{ResultsTable7}). \emph{Males} have 1.66 times higher odds (i.e., 66\% more likely) of preferring nervous movements than \emph{females}. Participants in \emph{North America} and \emph{Europe}, relative to participants in \emph{Africa}, are 56\% and 64\% less likely to prefer nervous movements compared to smooth movements. Male participants are accordingly more likely to prefer nervous movements, and contrary, participants from North America and Europe are more likely to prefer smooth movements.

Despite diverging likelihoods across gender and location, smooth movement remains the preferred condition considering all group-specific frequencies, corresponding, again, to the entire sample's preferences for smooth movements.

\begin{table}[ht]
\label{ResultsTable7}
\caption{Model parameters for MovType \emph{Smoothness}.}
\resizebox{\textwidth}{!}{%
\begin{tabular}{lllllll}
\hline
                  & \begin{tabular}[c]{@{}l@{}}Regression \\ Coefficient B\end{tabular} & \begin{tabular}[c]{@{}l@{}}Standard \\ Error\end{tabular} & Wald   & df & Sig.    & Exp(B) \\ \hline
Gender(Male)      & .504                                                                & .176                                                      & 8.174  & 1  & .004    & 1.655  \\ \hline
Location          &                                                                     &                                                           & 26.198 & 2  & $<$.001 &        \\ \hline
Location(Europe)  & -.814                                                               & .293                                                      & 7.704  & 1  & .006    & .443   \\ \hline
Location(Africa)  & -1.012                                                              & .198                                                      & 26.063 & 1  & $<$.001 & .363   \\ \hline
Experience(low)   & -.082                                                               & .175                                                      & .218   & 1  & .641    & .922   \\ \hline
Age(low)          & .080                                                                & .168                                                      & .230   & 1  & .631    & 1.084  \\ \hline
TechAffinity(low) & .127                                                                & .176                                                      & .523   & 1  & .470    & 1.135  \\ \hline
Education(low)    & .051                                                                & .174                                                      & .085   & 1  & .771    & 1.052  \\ \hline
Constant          & -.841                                                               & .244                                                      & 11.891 & 1  & $<$.001 & .431   \\ \hline
\end{tabular}%
}
\end{table}

\subsection{Speed}

A binary logistic regression analysis was conducted to predict preferences in the fifth video (condition B vs. A) based on gender, location, experience, age, technology affinity, and education. The overall model is statistically significant (${\chi}^2(7) = 31.00, p < .001$). The model explains between 3.3\% (Cox and Snell $R^2$) and 4.4\% (Nagelkerke $R^2$) of the variance in preference ratings and correctly classifies 59.6\% of cases.
\emph{Gender} ($b = .56, p < .001, OR = 1.75$) and \emph{age} ($b = .40, p = .004, OR = 1.49$) are significant predictors in the model (see Table~\ref{ResultsTable8}). \emph{Males}, relative to \emph{females}, have 1.75 higher odds of preferring fast movement over slow movement. With increasing \emph{age}, the odds of preferring fast movements decrease by 1.49 times.

Regarding group-specific frequencies, the influence of gender and age change the preferred speed condition in our study. While a slight majority of male participants (56.62\%) prefer the fast movement, 57.36\% of the female participants prefer the slow movement. Regarding age, participants 30 years old and below slightly prefer the fast movement (53.89\%), while 56.41\% of the participants who are 31 years old and above prefer the slow movement. Even though these differences are relatively small, the effect of gender and age on speed preferences represents a significant influence on speed preferences.

\begin{table}[ht]
\label{ResultsTable8}
\caption{Model parameters for MovType \emph{Speed}.}
\resizebox{\textwidth}{!}{%
\begin{tabular}{lllllll}
\hline
                  & \begin{tabular}[c]{@{}l@{}}Regression \\ Coefficient B\end{tabular} & \begin{tabular}[c]{@{}l@{}}Standard \\ Error\end{tabular} & Wald   & df & Sig.    & Exp(B) \\ \hline
Gender(Male)      & .561                                                                & .145                                                      & 15.010 & 1  & $<$.001 & 1.752  \\ \hline
Location          &                                                                     &                                                           & .162   & 2  & .922    &        \\ \hline
Location(Europe)  & .046                                                                & .254                                                      & .033   & 1  & .856    & 1.047  \\ \hline
Location(Africa)  & -.036                                                               & .179                                                      & .041   & 1  & .840    & .965   \\ \hline
Experience(low)   & -.166                                                               & .144                                                      & 1.322  & 1  & .250    & .847   \\ \hline
Age(low)          & .401                                                                & .138                                                      & 8.409  & 1  & .004    & 1.494  \\ \hline
TechAffinity(low) & -.055                                                               & .146                                                      & .143   & 1  & .705    & .946   \\ \hline
Education(low)    & .095                                                                & .146                                                      & .423   & 1  & .515    & 1.100  \\ \hline
Constant          & -.445                                                               & .215                                                      & 4.284  & 1  & .038    & .641   \\ \hline
\end{tabular}%
}
\end{table}

\section{Discussion}\label{Discussion}

To investigate the effect of individual differences on preferences for robot movements, we have presented a systematic literature review resulting in hypotheses that are based on existing findings. Next, we have conducted a Web-based experiment with 930 participants to evaluate the influence of demographic characteristics on five distinct MovTypes of an industrial robot: approach direction, movement range, rotation, smoothness and speed. For each MovType, GEE binary logistic regression and pairwise comparisons were carried out to evaluate the likelihood of preferring a condition per movement parameter, given the participants' demographic profiles. These analyses were complemented with group-specific frequencies to evaluate if the odds ratios found to be significant do actually change the majority preference per group.

Regarding the results presented in~\cite{Hostettler}, the overall preferences from our study correspond to their findings according to which humans perceive a large movement range, conventional rotations and smooth movements more human-like, which they have demonstrated to correlate with preference. Regarding approach direction and speed, no clear statement can be made based on their results, however, with our larger sample size we found a significant preference for the approach from the side and an equal distribution of preferences for slow and fast robot speed. Demographic characteristics do influence the preferences but do not change the majority-preferred condition for most MovTypes, in line with~\cite{Hostettler}, even though we found slight effects of age and gender for speed. As we have used the identical stimuli, differences might result from the smaller sample size in~\cite{Hostettler}.

In addition, our analyses allow to accept or reject the hypotheses derived from our literature review, as presented in Section~\ref{hypotheses}. 

Regarding \emph{Approach Direction}, we expected that males prefer an approach from the side, and that females prefer an approach straight towards the participant. All other demographic characteristics were not expected to influence human preferences, and no prediction was possible concerning the influence of technology affinity. Our results reveal no influence of gender, and both genders prefer the approach from the side, allowing us to accept the hypothesis for males, but to reject the hypothesis for females. Even though all groups prefer the approach from the side, we found significant differences between European and African participants as well as age differences that do, however, not change the majority's preferred condition. Prior experience with robots, technology affinity and educational level have no significant influence on approach direction preferences, and the corresponding hypotheses can therefore be accepted. The relevance of gender and prior experience for approach direction preferences found in~\cite{Syrdal} can, accordingly, not be supported. In line with~\cite{Syrdal}, academic background and technology affinity as possible approximation to computer proficiency do not influence approach direction preferences in our study. Possible explanations for the divergences to~\cite{Syrdal} can be found in differing robot appearances or inconsistencies in movement modulations between the two studies.

For \emph{Movement Range}, we expected that males and participants with more prior experience with robots prefer a small movement range, and that females prefer a large movement range. All other demographic characteristics were expected not to influence movement range preferences. Our results demonstrate that all groups prefer the large movement range, allowing us to confirm that age, technology affinity, location and educational level do not influence movement range preferences. Regarding our hypotheses, gender has no influence on the preferred movement range, and the hypothesis can, accordingly, be rejected. Experienced participants, on the other hand, show an even higher likelihood to prefer the large movement range, allowing us to reject our assumption. Contrary to~\cite{Story2022}, our results demonstrate an increased likelihood of preferring large movement range by participants with higher levels of experience with robots. However, participants with lower levels of experience also prefer the large movement range, and~\cite{Story2022} used other movement modulations as well as other dependent variables, and the results might therefore not be directly comparable. The absence of an effect of gender on movement range preferences is supported by the findings in~\cite{Leearticle, Bishop2019SocialAcceptance}, and likewise education does not influence movement range preferences, as in~\cite{Bishop2019SocialAcceptance}. On the other hand, \cite{Amir} found opposite-attraction effects according to which women (men) prefer larger (smaller) movement ranges. Our results do not support these findings based on the movement range modulations used.

Regarding MovType \emph{Rotation}, we expected that males prefer fewer rotations than females and that all other demographic characteristics do not influence preferences. Our results reveal that all groups prefer conventional rotations and preferences are not influenced by any demographic characteristic, verifying all hypotheses except for the influence of gender which can be rejected. \cite{Abel} found gender differences regarding perceived anthropomorphic movements which can be partly characterized by robot rotations. However, our results do not support an effect of gender on rotation preference. 

Regarding \emph{Smoothness}, we expected preferences for the smooth condition for younger, male, and experienced participants, whereas females were expected to prefer nervous movements. No effect was expected for location, technology affinity and educational level. As all groups clearly prefer the smooth movement, all hypotheses can be accepted, except for females who also prefer smooth movement. Still, male participants are even more likely to prefer nervous movements compared to females, and participants from North America and Europe are more likely to prefer smooth movements than African participants. Age, prior experience with robots, technology affinity, and education do not influence preferences. \cite{Abel} found gender to influence the sensitivity of perceiving robot movements as human-like. As they modulated digressive curves vs. point-to-point curves, their results are comparable to our smoothness preferences, even though modulation differences might affect comparability. Male participants are more likely to prefer nervous movements in our study, and our results therefore contradict the findings presented in~\cite{Abel}. 

For \emph{Speed}, we assumed that males and experienced participants prefer slow movements, and that females prefer fast movements. In addition, we have expected that age, location, technology affinity and educational level have no influence on speed preferences. Contrary to our expectations, our results reveal that males slightly prefer fast movements, compared to a slight female preference for slow movements, disconfirming our hypothesis. Participants of age 30 and below slightly prefer fast movements, and participants of age 31 and above slightly prefer slow movements. For gender as well as age, the likelihood of these specifications is significant. Speed preference are almost equally distributed in the entire sample, and group-specific analysis does not change the preferred condition. We therefore accept our hypotheses that technology affinity, educational level, experience with robots, and location have no significant influence. \cite{Brandl} investigated speed profiles and found no effect of age and gender, contradicting our result. This might possibly be a result of movement differences as they have used approach scenarios that might not be comparable to the robot movements used in our study. On the other hand, \cite{Story2022} found no effect of experience, and \cite{Vannucci} found no influence of participants' location on speed preferences. While their studies differ regarding robot types, samples and tasks, their findings are in line with the absent influence of experience and location on speed preferences.

In summary, the predictive power of static demographic characteristics is limited for the characteristics and MovTypes investigated in this study. Even though we found some indications of relevance for age, gender, experience with robots, and location, the group differences mostly are not strong enough to cause a difference in majority-preferred conditions. Our comparison with existing findings reveals some agreements, but also several deviations that can possibly be explained through a variety of reasons, such as different robot types used, inconsistent movement modulations, different DVs as well as very heterogeneous sample structures and sizes. 

To achieve improved acceptance and psychological safety, and to reduce cognitive workload when working with robots, movement characteristics of industrial robots have been demonstrated to influence human perception. As industrial robots are mostly deployed in organizations to ensure manufacturing companies' competitive position, they need to increase operational effectiveness. It is therefore important to align human-related goals such as increased acceptance and psychological safety with the organizational goals of efficiency and effectiveness. It is noteworthy in this context that most of the MovTypes investigated allow to match human preferences while considering organizational goals, even though there might be tradeoffs for certain movement adjustments. Speed reduction for instance leads to lower task cycle times, but might increase operator well-being. Movement range, on the other hand, has to be adjusted to the functional requirement of the robot cell, limiting the room for adjustments. However, smoothness of movements can be modulated without affecting organizational goals. Our results therefore represent a movement-related approach that considers several human/operator-related desires that can be used without having negative effects on organizational goals. 

\section{Conclusion}
\label{Conclusion}

We have conducted a Web-based experiment with 930 participants and analysed the effect of gender, age, location, technology affinity experience with robots and education on the perception of five MovTypes of an industrial robot: approach direction, movement range, rotation, smoothness and speed. With our study, we aim to clarify if and how demographic characteristics influence the perception of movements of an industrial robot. Our results reveal that humans prefer an approach from the side over a straight approach towards them, a large movement range over a small movement range, conventional rotations over unconventional rotations, smooth movements over nervous movements, and an almost equal preference for slow or fast speeds. Regarding individual differences, most of these preferences are robust to demographic variation, and only gender and age were found to cause slight preference differences between slow and fast movements. The used movement modulations allow to evaluate human preferences for distinct MovTypes based on the work presented in~\cite{Hostettler}, and represents, to the best of our knowledge, one of the first investigations of the effects of individual differences on such selective movement characteristics of an industrial robot so far. 

\subsection{Limitations}
Our approach comes with several limitations that need to be considered. Our Web-based experiment only allows to analyze the effect on robot movement observation which might not be considered an actual interaction or even a collaboration with robots. However, observing a moving robot corresponds to the case of co-existence which is already reality in many shopfloors where robots work behind fences or in production lines. Furthermore, the Web-based nature of our experiment creates a distance between the observer and the robot, and possibly reduces the impressiveness of certain robot movements. On the other hand, our approach allows to carry out a study with a large sample size.
Second, we have used existing movement patterns that are clearly distinctive and irrespective of actual tasks a robot may need to perform, and movements in actual implementations might be more complex and interconnected. This ensures a clear evaluation of these separate movement characteristics without possibly confounding effects. Moreover, these simplified movement patterns are explicitly definable and can be used for almost any motion sequence an actual task may require.
Regarding our experimental setup, we have only used one robot type: the six DoF UR 10e. Even though most articulated robots with six DoF have a similar appearance, they might not appear similar to human observers. In addition, we have transferred certain movement patterns from humanoid robots to an articulated robot, which might not necessarily work in the opposite direction, which limits the generalizability of our findings for humanoid robots. However, the underrepresentation of research on industrial robots in HRI as well as their wide diffusion in shopfloors calls for results that directly relate to these non-humanoid devices. 
Last, even though we have ensured a sufficiently large sample size to analyse the influence of individual differences, we had to merge some sub-groups due to underrepresentation of certain characteristics, and we have completely ignored some groups as they have not reached an analyzable sample size. In addition, approximately 40\% of the sample had never experienced or interacted with robots, and an expert sample with humans who actually interact with and use robots regularly would increase the results' generalizability. Still, we believe that the effects of individual differences when observing robots happen very directly and unconsciously, and that the variety of individual differences leads to preference differences within an expert sample as well.

\subsection{Outlook}
To further develop the understanding of robot movement perception and to design interactions that promote industrial robot acceptance, we propose several future research directions and experiments. Responding to our limitations, investigations of physical interactions with robots as well as with other robot types and actual robot users are required. Web-based approaches might want to replicate our study with a sample that includes sufficient sample sizes for groups that we have not considered in our analysis. 

With regard to future directions and as we have found human perception to be rather robust to static demographic variation, we call for investigations of more dynamic, behavioral data such as proximity behavior and physiological variables such as stress-related characteristics like gaze behavior, pupil dilation, or heart rate variability that might explain differing perceptions of robots and their movements, and how these variations affect additional constructs such as cognitive workload or psychological safety. Moreover, future research might permit combining human-related considerations with organizational goals, allowing to balance various relevant requirements. With regard to the nature of behavioral and physiological data streams, they not only have the potential to provide more information in terms of which characteristics drive robot acceptance, but they are also collectable at run time compared to the rather time-consuming completion and analysis of questionnaires. Findings on such data have the potential to build the foundation for future run-time adaptive and reactive robotic systems, enabling autonomous acceptance-promoting robotic systems in HRI, and in particular for prospective HRC.

\section*{Declarations}\label{sec8}
\subsection*{Data Availability}
The collected and analysed data of this study is available in a public OSF data repository,  \url{https://osf.io/ht2qj/?view_only=bb549d07a92e4822be21c7e21571b965}.

\subsection*{Funding and Competing Interests}

This research was partially funded by the Chair for Interaction- and Communication-based Systems at the Institute of Computer Science, University of St. Gallen, Switzerland.

\subsection*{Ethical Approval}

The Ethics Committee of the University of St.Gallen has confirmed that no ethical approval is required.

\bibliography{sn-article}

\newpage
\appendix
\section{Appendix}
\label{AppendixA}

\begin{table}[ht]

\caption{Group-specific frequencies per MovType.}
\label{group-specificFrequencies}
\resizebox{\textwidth}{!}{%
\begin{tabular}{llllll}
            & \textbf{ApproachDir} & \textbf{MovRange} & \textbf{Rotation} & \textbf{Smoothness} & \textbf{Speed} \\ \cline{2-6} 
            & \multicolumn{5}{c}{\textbf{Male}}                                                                   \\ \hline
Condition A & 59.00\%              & 39.70\%           & 64.64\%           & 75.27\%             & 43.38\%        \\
Condition B & 41.00\%              & 60.30\%           & 35.36\%           & 24.73\%             & 56.62\%        \\
            & \multicolumn{5}{c}{\textbf{Female}}                                                                 \\ \hline
Condition A & 62.26\%              & 41.15\%           & 63.11\%           & 79.74\%             & 57.36\%        \\
Condition B & 37.74\%              & 58.85\%           & 36.89\%           & 20.26\%             & 42.64\%        \\
            & \multicolumn{5}{c}{\textbf{30 and below}}                                                           \\ \hline
Condition A & 57.22\%              & 42.59\%           & 63.70\%           & 76.11\%             & 46.11\%        \\
Condition B & 42.78\%              & 57.41\%           & 36.30\%           & 23.89\%             & 53.89\%        \\
            & \multicolumn{5}{c}{\textbf{31 and above}}                                                           \\ \hline
Condition A & 65.38\%              & 37.44\%           & 64.10\%           & 79.49\%             & 56.41\%        \\
Condition B & 34.62\%              & 62.56\%           & 35.90\%           & 20.51\%             & 43.59\%        \\
            & \multicolumn{5}{c}{\textbf{Lower educational level}}                                                \\ \hline
Condition A & 60.82\%              & 43.30\%           & 63.57\%           & 76.63\%             & 47.42\%        \\
Condition B & 39.18\%              & 56.70\%           & 36.43\%           & 23.37\%             & 52.58\%        \\
            & \multicolumn{5}{c}{\textbf{Higher educational level}}                                               \\ \hline
Condition A & 60.56\%              & 39.12\%           & 64.01\%           & 77.93\%             & 51.80\%        \\
Condition B & 39.44\%              & 60.88\%           & 35.99\%           & 22.07\%             & 48.20\%        \\
            & \multicolumn{5}{c}{\textbf{No experience at all}}                                                   \\ \hline
Condition A & 63.86\%              & 44.57\%           & 65.76\%           & 79.35\%             & 53.80\%        \\
Condition B & 36.14\%              & 55.43\%           & 34.24\%           & 20.65\%             & 46.20\%        \\
            & \multicolumn{5}{c}{\textbf{Higher level of experience}}                                             \\ \hline
Condition A & 58.54\%              & 37.72\%           & 62.63\%           & 76.33\%             & 48.22\%        \\
Condition B & 41.46\%              & 62.28\%           & 37.37\%           & 23.67\%             & 51.78\%        \\
            & \multicolumn{5}{c}{\textbf{North America}}                                                          \\ \hline
Condition A & 58.65\%              & 40.38\%           & 63.46\%           & 78.85\%             & 50.00\%        \\
Condition B & 41.35\%              & 59.62\%           & 36.54\%           & 21.15\%             & 50.00\%        \\
            & \multicolumn{5}{c}{\textbf{Europe}}                                                                 \\ \hline
Condition A & 62.89\%              & 41.15\%           & 65.53\%           & 81.21\%             & 50.31\%        \\
Condition B & 37.11\%              & 58.85\%           & 34.47\%           & 18.79\%             & 49.69\%        \\
            & \multicolumn{5}{c}{\textbf{Africa}}                                                                 \\ \hline
Condition A & 53.85\%              & 37.91\%           & 58.24\%           & 63.74\%             & 51.10\%        \\
Condition B & 46.15\%              & 62.09\%           & 41.76\%           & 36.26\%             & 48.90\%        \\
            & \multicolumn{5}{c}{\textbf{Lower technology affinity}}                                              \\ \hline
Condition A & 61.73\%              & 40.59\%           & 63.00\%           & 77.17\%             & 53.91\%        \\
Condition B & 38.27\%              & 59.41\%           & 37.00\%           & 22.83\%             & 46.09\%        \\
            & \multicolumn{5}{c}{\textbf{Higher technology affinity}}                                             \\ \hline
Condition A & 59.52\%              & 40.26\%           & 64.77\%           & 77.90\%             & 46.83\%        \\
Condition B & 40.48\%              & 59.74\%           & 35.23\%           & 22.10\%             & 53.17\%       
\end{tabular}%
}
\end{table}



\end{document}